\documentclass{article}

\usepackage[round]{natbib}

\usepackage[utf8]{inputenc} 
\usepackage[T1]{fontenc}    
\usepackage[colorlinks=true,citecolor=blue,linkcolor=orange]{hyperref}       
\usepackage{url}            
\usepackage{booktabs}       
\usepackage{amsfonts}       
\usepackage{nicefrac}       
\usepackage{microtype}      
\usepackage{xcolor}         
\usepackage{xspace}

\usepackage{amssymb,amsmath}
\usepackage{amsthm}
\usepackage{cleveref}
\usepackage{bm}
\usepackage{mathtools}
\usepackage{todonotes}
\usepackage{float}

\newtheorem{theorem}{Theorem}

\graphicspath{{img/}}

\newcommand{\E}{\mathbb{E}}
\renewcommand{\Pr}{\mathbb{P}}

\DeclareMathOperator{\kl}{kl}

\allowdisplaybreaks    
\usepackage[
    verbose=true,
    letterpaper,
    bottom=1in,
    left=3cm,
    right=3cm,
    top=1in,
    headheight=12pt,
    headsep=25pt,
    footskip=30pt]{geometry}

\makeatletter
    \def\@maketitle{%
  \newpage
  \null
  \vskip 1em%
  \begin{center}%
  \let \footnote \thanks
    {\LARGE \@title \par}%
    \vskip 1.5em%
    {\small
      \lineskip .5em%
      \begin{tabular}[t]{c}%
        \@author
      \end{tabular}\par}%
  \end{center}%
  \par
  \vskip 0.1em}
\makeatother

\title{A Note on the Efficient Evaluation of PAC-Bayes Bounds}

\author{%
  Felix Biggs\footnote{Correspondence to: \texttt{contact@felixbiggs.com}} \\
  Department of Computer Science\\
  University College London and Inria\\
  London \\
}

\begin{document}

\maketitle

\begin{abstract}
  When utilising PAC-Bayes theory for risk certification, it is usually necessary to estimate and bound the Gibbs risk of the PAC-Bayes posterior.
  Many works in the literature employ a method for this which requires a large number of passes of the dataset, incurring high computational cost.
  This manuscript presents a very general alternative which makes computational savings on the order of the dataset size.
\end{abstract}



\section{Overview}

When evaluating PAC-Bayes bounds it is usually necessary to calculate the in-sample Gibbs risk of the PAC-Bayes posterior, \(\rho\).
For a fixed dataset, \(s = \{z_1, \dots, z_m\}\), the in-sample risk is the average loss, \(L_s(h) = \frac{1}{m} \sum_{i=1}^m \ell(h, z_i)\), for loss \(\ell \in [0, 1]\).\footnote{Extensions to arbitrary bounded losses are also possible by re-scaling.}
This is generalised in PAC-Bayes by the the Gibbs loss (for which we slightly abuse notation), \(L_s(\rho) := \mathbb{E}_{H \sim \rho} L_s(H)\).

For many choices of posterior this is not directly tractable, so a Monte Carlo estimate is made from posterior samples.
For true ``risk certification'' in the final PAC-Bayes bound, a high-probability upper bound on \(L_s(\rho)\) is needed, generally using this estimate.
The most common formulation seen in the literature (for example, in \citealp{DBLP:conf/nips/LangfordC01,dziugaite2017computing,DBLP:conf/aistats/ClericoDD22,DBLP:journals/corr/abs-2111-07737}, and many more works), fully evaluates \(L_s(H)\) for \(n\) samples of \(H\) from the posterior, requiring \(n\) passes over the entire dataset.
The following\footnote{This is actually a refinement of the result given in \citet{DBLP:conf/nips/LangfordC01}, which removes a factor of \(\log 2\) and is valid for \(\ell \in [0, 1]\) rather than \(\ell \in \{0, 1\}\).} can be used to obtain the statistical guarantee needed.

\begin{theorem}[\citealp{DBLP:conf/nips/LangfordC01}]\label{th:old}
  For fixed dataset \(s\), posterior \(\rho\), \(n \in \mathbb{N}\), bounded loss \(\ell \in [0, 1]\), \(\delta \in (0, 1)\), and i.i.d. posterior samples \(H_1, \dots, H_n\) from \(\rho\), with probability at least \(1{-}\delta\),
  \[ L_s(\rho) \le \kl^{-1}\left(\frac{1}{n} \sum_{i=1}^n L_s(H_i), \, \frac{1}{n} \log\frac{1}{\delta}\right).\]
  Here \(\kl^{-1}(q, c) := \sup \{p \in [0, 1] : \kl(q, p) \le c\}\) is a generalised inverse of \(\kl(q, p) := q \log\frac{q}{p} + (1-q) \log\frac{1-q}{1-p}\), the Kullback-Lieber divergence between Bernoulli distributions with parameters \(q\) and \(p\).
\end{theorem}

Since \(\kl^{-1}(q, c) \le q + \sqrt{c/2}\), the gap between Monte Carlo estimate and true value converges at rate at least \(O(n^{-\frac12})\), but this bound is considerably sharper when the Gibbs loss is close to zero (the realisable case).
The number of passes over the dataset required can however be very large (in \citealp{dziugaite2017computing}, \(n{=}150000\)), making this method very computationally expensive, potentially with a cost of many GPU hours \citep{DBLP:conf/aistats/ClericoDD22}.

\paragraph{Alternative method.}
An alternative approach uses a new posterior sample for each dataset example, while still looping through the dataset \(n\) times.
Surprisingly, this is a valid approach, and the statistical guarantee is considerably stronger, so that \(O(m)\) fewer passes over the dataset are necessary for the same guarantee.
The computational savings of this method may then also be \(O(m)\).

\begin{theorem}[Main Result]\label{th:new}
  For fixed dataset \(s\), posterior \(\rho\), \(n \in \mathbb{N}\), bounded loss \(\ell \in [0, 1]\), \(\delta \in (0, 1)\) and i.i.d. posterior samples \(H_1, \dots, H_{nm}\) from \(\rho\), with probability at least \(1{-}\delta\),
  \[ L_s(\rho) \le \kl^{-1}\left(\frac{1}{nm} \sum_{i=1}^n \sum_{j=1}^m \ell(H_{j + m(i-1)}, z_j), \, \frac{1}{nm} \log\frac{1}{\delta}\right).\]
\end{theorem}

A basic version of this method with a weaker statistical guarantee is given in \citet[][Theorem 5.1; their result is implied by ours through the \(\operatorname{kl}\) relaxation above]{pmlr-v162-biggs22a}, and may appear further back in the literature, but it appears little-known and is not the primary focus of that work.
\Cref{th:new} is proved in the next section, and an extension to test set bounds is given in \Cref{section:extension}.

\section{Proof of \Cref{th:new}}

\Cref{th:new} is essentially implied by the following generalisation of the Chernoff inequality (\Cref{th:hoeffext}) and an inverted version of it (\Cref{th:inverse}).
The difference of this result from the form used in proving \Cref{th:old} is that the means of the summed random variables are allowed to differ.

\begin{theorem}[Hoeffding extension, \citet{hoeffding63}]\label{th:hoeffext}
    Let \(X_1, . . . , X_T\) be independent random variables with \(X_i \in [0, 1]\) and \(\E[X_i] = p_i\).
    Define \(X := T^{-1} \sum^T_{i=1} X_i\) and \(p := \E[X] = T^{-1} \sum^T_{i=1} p_i\). For any \(t \in [0, p]\),
    \[ \Pr(X \le t) \le e^{- T \kl(t, p)}. \]
\end{theorem}

\begin{proof}
  The result is proved through the Cramer-Chernoff method.
  The moment generating function (MGF) of each independent variable is bounded by a Bernoulli MGF through the observation that \(t \mapsto e^{\lambda t}\) is convex, and the product of MGFs is bounded by the MGF for a Binomial, with parameters \((T, p)\), through the arithmetic-geometric mean inequality.
\end{proof}

A corollary of this theorem is the following inverted formulation.
This result can be used to prove both \Cref{th:old} and \Cref{th:new}.

\begin{theorem}[Hoeffding extension inverse]\label{th:inverse}
    Let \(X_1, . . . , X_T\) be independent random variables with \(X_i \in [0, 1]\) and \(\E[X_i] = p_i\).
    Define \(X := T^{-1} \sum^T_{i=1} X_i\) and \(p := \E[X] = T^{-1} \sum^T_{i=1} p_i\).
    Then, for any \(\delta \in (0, 1)\), with probability at least \(1{-}\delta\),
    \[ p \le \kl^{-1}\left(X, \, \frac{1}{T} \log\frac{1}{\delta}\right).\]
\end{theorem}

\begin{proof}
  Firstly, define a one-sided version of the small kl (as is used in \citet{langford05} to obtain bounds containing \(\log(1/\delta)\) rather than \(\log(2/\delta)\)):
  \[
    \kl_+(q, p) :=
    \begin{cases}
      \kl(q, p) & \text{for } q \le p, \\
      0 & \text{else}
    \end{cases}.
  \]
  This function has two important properties, holding for any choices of the variables:
  \begin{align*}
  \kl_+(x, p) > \kl_+(y, p)   &\implies   x < y \implies x \le y, \\
    p \le \kl^{-1}(q, c) &\iff \kl_+(q, p) \le c.
  \end{align*}
  The first of these follows as the contrapositive of the definition of a non-increasing function, since \(\kl_+(q, p)\) is non-increasing as a function of \(q\).
  The second follows since
  \[ \kl^{-1}(q, c) = \sup \{p \in [0, 1] : \kl(q, p) \le c\} = \sup \{p \in [0, 1] : \kl_+(q, p) \le c\},\]
  as the supremum will always be in the right hand part of the function where \(p \ge q\).

  Defining \(p_i, p\) and \(X\) as in \Cref{th:hoeffext}, with \(t \in [0, p]\)
  \begin{align*}
    \Pr\left( \kl_+(X, p) > \kl_+(t, p) \right) \le \Pr\left( X \le t \right) \le e^{- T\kl(t, p)} = e^{- T\kl_+(t, p)}.
  \end{align*}
  These steps follow by applying the first property of \(\kl_+\), \Cref{th:hoeffext}, and the equivalence of \(\kl\) and \(\kl_+\) on this part of the domain.
  This result further implies that for any \(c > 0\),
  \[ \Pr\left( \kl_+(X, p) > c \right) \le e^{-nc},\]
  since if \(c\) is larger than \(\kl_+(0, p)\), the probability of the event is zero.

  The proof is completed by taking the complement of this statement with \(c = T^{-1} \log(1/\delta)\), and applying the second property of \(\kl_+\).
\end{proof}

\begin{proof}[Proof of \Cref{th:new}]
  For samples \(H_1, \dots, H_{nm}\), define \(X_{j + m(i-1)} := \ell(H_{j + m(i-1)}, z_j)\) and \(p_t := \E X_t\).
  As in \Cref{th:inverse}, define \(X = \frac{1}{nm} \sum_{t =1}^{nm} X_t = \frac{1}{nm} \sum_{i=1}^n \sum_{j=1}^m \ell(H_{j + m(i-1)}, z_j)\), and
  \[ p := \frac{1}{nm} \sum_{t=1}^{nm} p_t = \frac{1}{nm} \sum_{i=1}^n \sum_{j=1}^m \mathbb{E}_{H \sim \rho} [\ell(H, z_j)] =  \mathbb{E}_{H \sim \rho} \left[\frac{1}{nm} \sum_{i=1}^n \sum_{j=1}^m \ell(H, z_j) \right] = L_s(\rho), \]
  which follows by the linearity of the expectation.
  The result then follows from \Cref{th:inverse} with \(T = nm\).
\end{proof}

\section{Test Sets and Large Datasets}\label{section:extension}

A slight variation of \Cref{th:new} can be also be used to obtain test set bounds on the out-of-sample Gibbs risk, \(L(\rho) := \E_{S \sim \mathcal{D}^m} L_S(\rho)\), where \(\mathcal{D}\) is the data-generating distribution.

\begin{theorem}[Test Set Gibbs Bound]
  For fixed distribution \(\mathcal{D}\), posterior \(\rho\), \(m \in \mathbb{N}, n \in \mathbb{N}\), bounded loss \(\ell \in [0, 1]\), \(\delta \in (0, 1)\), draw i.i.d. posterior samples \(H_1, \dots, H_{nm}\) from \(\rho\), and samples \(Z_1, \dots, Z_m\) from \(\mathcal{D}\).
  With probability at least \(1{-}\delta\),
  \[ L(\rho) \le \kl^{-1}\left(\frac{1}{nm} \sum_{i=1}^n \sum_{j=1}^m \ell(H_{j + m(i-1)}, Z_j), \, \frac{1}{nm} \log\frac{1}{\delta}\right).\]
\end{theorem}

\begin{proof}
  The proof is essentially the same as that of \Cref{th:new}, except for the changed definition \(X_{j + m(i-1)} = \ell(H_{j + m(i-1)}, Z_j)\) using the randomised samples \(Z_j\), which has mean \(\E X_i = L(\rho)\).
\end{proof}

For extremely large datasets, even cheaper statistical guarantees could be obtained by sub-sampling using the test set theorem with \(\mathcal{D} = \operatorname{Uniform}(s)\), or by application of each \(H_t\) to a mini-batch (which could be computationally cheaper to implement in stochastic deep networks).


\nocite{DBLP:journals/corr/abs-2205-07880}

{
\small
\bibliography{bibliography}
}

\end{document}